%% file: rvc2020.tex
\documentclass[twocolumn,9pt]{extarticle}

\usepackage[utf8]{inputenc}
\usepackage{amsmath,amssymb} 

\usepackage{comment}
\usepackage{hyperref}

\input{math}
\usepackage[a4paper,left=1.5cm,right=1.5cm,top=1.7cm,bottom=1.7cm,footskip=.25in]{geometry}

\begin{document}

\title{\huge Multi-domain semantic segmentation
   with pyramidal fusion\thanks{The authors wish 
   to thank Ivan Krešo for useful discussions.
   This work has been supported by
   European Regional Development Fund,
   grants
   KK.01.1.1.01.0009,
   KK.01.2.1.01.0022, and
   KK.01.2.1.01.0128.
   This work has also been supported by
   Microblink ltd and Rimac automobili ltd,
   as well as by 
   College for Information Technologies
   from Zagreb,
   which provided access
   to 6 GPU Tesla V100 32GB.
   }
   }

\author{Marin Oršić, Petra Bevandić, 
    Ivan Grubišić,
    Josip Šarić, and Siniša Šegvić}
\date{University 
  of Zagreb
  Faculty of Electrical Engineering and Computing}

\maketitle
\thispagestyle{empty}

\begin{abstract}
  We present our submission to
  the semantic segmentation contest
  of the Robust Vision Challenge
  held at ECCV 2020.
  The contest requires 
  submitting the same model
  to seven benchmarks
  from three different domains.
  Our approach is based on 
  the SwiftNet architecture 
  with pyramidal fusion.
  We address inconsistent taxonomies
  with a single-level 
  193-dimensional 
  softmax output.
  We strive to train with 
  large batches in order 
  to stabilize optimization
  of a hard recognition problem,
  and to favour smooth evolution 
  of batchnorm statistics.
  We achieve this by implementing 
  a custom backward step
  through log-sum-prob loss, 
  and by using small crops
  before freezing the population statistics.
  Our model ranks first on the RVC 
  semantic segmentation challenge
  as well as on the WildDash 2 leaderboard.
  This suggests that pyramidal fusion
  is competitive not only 
  for efficient inference
  with lightweight backbones,  
  but also in large-scale setups
  for multi-domain application.
\end{abstract}

\section{Introduction}

Large realistic datasets
\cite{cordts16cvpr,neuhold17iccv}
have immensely contributed 
to the development of techniques 
for semantic segmentation 
of road-driving scenes. 
The reported accuracies grew rapidly 
over the last five years. 
However the researchers soon realized
that the learned models 
were often doing poorly in the wild. 
Hence there emerged a desire 
to design models which 
simultaneously address 
multiple datasets from several domains
\cite{kreso18arxiv,lambert20cvpr}, 
akin to combined events in athletics. 
The second Robust Vision Challenge 
proposes a contest 
over 7 semantic segmentation datasets
across three domains: 
interiors, photographs, road-driving.
Each participant has to submit 
the same model to all 7 datasets.
This report presents 
the main insights
which we gathered 
while participating 
in the challenge.

\section{Datasets}

Besides knowledge transfer
from ImageNet,
we train exclusively
on the seven datasets 
of the challenge
which we summarize in Table~\ref{tab:datasets}. 
ADE20k contains rather small 
indoor and outdoor images,
and has the most classes.
ScanNet contains interior images
with very noisy labels.
Cityscapes, KITTI, Vistas, VIPER, 
and WildDash 2 contain road-driving images.
Cityscapes contains images 
from western Europe
taken with the same camera 
in fine weather.
Vistas contains crowdsourced images
across the globe in all kinds of weather.
WildDash 2 collects hand-picked images
according to a system of hazards
\cite{zendel18eccv}.
VIPER contains images generated
by a computer game.
However, its labeling is inconsistent 
with the remaining road-driving datasets.
E.g.\ a person seen through a windscreen
is labeled as 
class person 
in VIPER,
while other road-driving datasets
label such pixels with 
a suitable vehicle class.

\begin{table}[htb]
\centering
\begin{tabular}{llrr@{ - }lr@{$\pm$}l}
Dataset & content 
  & \multicolumn{1}{c}{size}
  & \multicolumn{2}{c}{class count}
  & \multicolumn{2}{c}{resolution} 
  \\
\hline
ADE20K       & photos   &  22210
  & 150 & 150& 460  & 154\\
Cityscapes   & driving  &   3475 
  & 28 & 19  & 1448 & 0   \\
KITTI        & driving  &    200 
  & 28 & 19  & 682  & 1   \\
VIPER        & artificial &  18326 
  & 32 & 19  & 1440 & 0  \\
ScanNet      & interior &  24902 
  & 40 & 20  & 1109 & 78  \\
Vistas       & driving  &  20000 
  &  65 & 65 & 2908 & 608 \\
WildDash 2     & driving  &   4256 
  & 26 & 20  & 1440 & 0   
\end{tabular}
\caption{Dataset summary.
  Size denotes 
  the total number of annotated 
  non-test images.
  Class count denotes 
  the total number of training 
  and test classes.
  Resolution denotes
  the mean and standard deviation 
  of the square root 
  of the number of pixels
  ($\sqrt{HW}$) 
  across the training split.
  }
\label{tab:datasets}
\end{table}


\section{Universal taxonomy}

We consider semantic classes 
as sets of pixels 
in all possible images 
and use set notation 
to express relations between them.
%
We build a flat universal taxonomy
defined by mappings 
from dataset-specific classes
to subsets of a universal set 
of disjoint elementary classes
\cite{lambert20cvpr}.
The taxonomy can be produced
by iterative application 
of the following rules.
Note that WD and CS stand for 
WildDash and Cityscapes, 
respectively.
\begin{enumerate}
\itemsep0em
\item 
  If a class exactly matches another class,
  they are merged. 
  \\\emph{Example}: 
    $\mathrm{WD\text-sky} = 
      \mathrm{CS\text-sky}$ results in 
    dataset-specific mappings
    $\mathrm{WD\text-sky}\mapsto
      \cbr{\mathrm{sky}}$ 
    and 
    $\mathrm{CS\text-sky}\mapsto
      \cbr{\mathrm{sky}}$.
\item
  If a class is a subset
  of another class,  
  $c_i\subset c_j$, 
  the superset class is replaced 
  with the difference:
  $c_j'=c_j\setminus c_i$.
  Dataset mappings are updated 
  by replacing $c_j$ with 
  $\cbr{c_j', c_i}$.
  \\\emph{Example} (abstraction): 
  $\mathrm{WD\text-van}\subset\mathrm{KITTI\text-car}$
  gives %
  $\mathrm{KITTI\text-car} 
    \mapsto \cbr{\mathrm{car}, \mathrm{van}}$ 
  and 
  $\mathrm{WD\text-van}
    \mapsto\cbr{\mathrm{van}}$.
  \\\emph{Example} (abstraction): 
  $\mathrm{Vistas\text-bicyclist}\subset
    \mathrm{CS\text-rider}$
  results in 
  $\mathrm{CS\text-rider} 
    \mapsto \cbr{\mathrm{bicyclist}, 
      \mathrm{motorcyclist}}$ 
  and 
  $\mathrm{Vistas\text-bicyclist}
    \mapsto\cbr{\mathrm{bicyclist}}$. 
  \\\emph{Example} (composition): 
  $\mathrm{Vistas\text-pothole} 
    \subset\mathrm{VIPER\text-road}$
  results in 
  $\mathrm{VIPER\text-road} 
    \mapsto 
    \cbr{\mathrm{road\_other}, \mathrm{pothole}}$ 
  and 
  $\mathrm{Vistas\text-pothole}
    \mapsto\cbr{\mathrm{pothole}}$.
\item
If two classes are overlapping, 
$c_i \setminus c_j \neq \emptyset
 \;\wedge\;
 c_j \setminus c_i \neq \emptyset$, 
they are split into $c_i' = c_i\setminus c_j$, $c_j' = c_j\setminus c_i$, and $c=c_i\cap c_j$, so that there is no overlap. Dataset-specific mappings are updated by replacing $c_i$ with $\cbr{c_i', c}$ and $c_j$ with $\cbr{c_j', c}$.
\\\emph{Example}:
$\mathrm{VIPER\text-truck} 
  = \mathrm{truck} \cup \mathrm{pickup}$ and $\mathrm{ADE20K\text-truck} 
  = \mathrm{truck} \cup \mathrm{trailer}$
results in $\mathrm{VIPER\text-truck} \mapsto \cbr{\mathrm{truck}, \mathrm{pickup}}$ and $\mathrm{ADE20K\text-truck}\to\cbr{\mathrm{truck}, \mathrm{trailer}}$
(note that truck, pickup and trailer
   are disjoint).
\end{enumerate}
However, there are cases 
where applying the rule 3 
would make the taxonomy too complex. 
For instance, Vistas labels 
vehicle windows as vehicles, 
but VIPER labels these pixels
with what is seen behind. 
We ignore the overlap by making
simplifying assumptions such as 
$\mathrm{VIPER\text-car}=
 \mathrm{Vistas\text-car}$ and
 $\mathrm{Vistas\text-car}
  \;\cap\; \mathrm{VIPER\text-person}=
  \emptyset$.
This particular issue 
can not be properly resolved
without relabeling.
In practice, we first fuse Vistas
(road-driving dataset 
 with the finest granularity)
and ADE20K 
(the dataset with most classes).
Subsequently, we add classes 
from other datasets 
which require finer granularity. 
This results in one-to-many mappings
from each dataset to the universal set 
of 193 elementary classes.
We provide the source code 
for mapping dataset classes 
to the universal
taxonomy\footnote{\scalebox{.9}{\url{https://drive.google.com/drive/folders/1Wi4Uku2ERaciLAVlUKCXVCozZWooZ27L}}}.

\section{Method}

Our convolutional model
receives a colour image
and produces dense predictions
into 193 universal classes.
We use a SwiftNet architecture
with pyramidal fusion \cite{orsic20pr}.
We apply a shared ResNet-152 backbone
at three levels 
of a Gaussian resolution pyramid
and use 256 feature maps
along the upsampling path.
Our predictions are 8 times subsampled
with respect to the input resolution
in order to decrease 
the memory footprint.
We produce the logits at full resolution 
by 8$\times$ bilinear upsampling,
and recover predictions 
by summing softmax probabilities 
of all universal classes
which map to the particular dataset class.

We recover the probability 
of the void class 
by summing probabilities
of all universal classes
which do not map 
to the particular dataset.
We obtain crisp predictions 
by applying argmax over
all test classes, 
except on WildDash 2
where we apply the argmax
over all test classes
plus the void class.

We train the model with a compound loss 
which modulates the negative 
log-likelihood
in order to prioritize 
poorly classified pixels
and pixels at boundaries
\cite{orsic20pr}.
We avoid caching
logits at full resolution
by implementing the log-sum-prob loss
as a layer with a custom backprop step.
This decreases the memory footprint
during training by 2.54 GB per GPU.

\section{Training details}

We train our submission on
6 Tesla V100 GPUs with 32GB RAM.
We use random horizontal flipping, 
scale jittering and square cropping 
according to the schedule 
from Table \ref{tab:training}.
We attempt to alleviate 
noisy ScanNet labels
by setting the boundary modulation
to 1 (minimum) for all ScanNet crops.

We optimize our model with Adam.
The learning rate is attenuated 
by cosine annealing 
from $10^{-3}$ to $6\cdot10^{-6}$.
We freeze all batch normalization layers
at epoch 50 and train for 3 more epochs.
Once we freeze the batchnorm, 
we reset the gradient moments 
used by Adam.
The training involves 140k iterations, 
which took around 4 days on our hardware.

We sample crops from
93,369 training images 
from all datasets.
We favour fair representation
of classes and datasets
by composing mini-batches with
roulette wheel sampling.
In particular, we encourage sampling
of images with multiple class instances
and images with rare classes.


The biggest challenges were
the sheer extent of training data
and the details of 
multi-GPU implementation.
We do not use batchnorm syncronization
for simplicity and speed of training.
Accordingly, the population statistics
are updated on only one GPU.
Conversely, we perform model updates
by accumulating gradients from all GPUs.
We did not find enough time
to implement gradient checkpointing  
in the multithreaded environment.
Hence, we based our solution
on a backbone which provides
a reasonable performance
without checkpointing.
We did not use photometric jittering
since we were not sure that
our model has enough capacity
for such recognition problem.
We have had to avoid 
training on unlabeled images 
with unsupervised loss
in order to meet the deadline.

\begin{table}[htb]
\centering
\begin{tabular}{crcrr}
Epochs    & crop size & batch size 
  & jitter range & speed \\
\hline
0 -- 15   & 384          & 6$\times$16 
  & 0.75 -- 1.33       & 45 fps
\\
16 -- 31  & 512          & 6$\times$8   
  & 0.60 -- 1.67        & 27 fps         
\\
32 -- 49  & 768          & 6$\times$4    
  & 0.50 -- 2.00         & 14 fps         \\
50 -- 52  & 1024         & 6$\times$2 
  & 0.40 -- 2.50         & 9 fps 
\\
\end{tabular}
\caption{
  Mini-batch configuration schedule
    across the training epochs.
    The columns show square crop size,
    batch size, and the
    range of uniform scale jittering.
    The final column shows 
    how many crops are processed 
    in each second of training.
}
\label{tab:training}
\end{table}

\section{Results}

We evaluate all test images
on only three scales 
due to limited time.
Pixels where argmax 
correponds to the void class
occur on books (ScanNet), 
railroad (KITTI),
cobblestone (Cityscapes) etc.
We find most such pixels
in ScanNet (9.97\%),
Cityscapes (7.84\%), 
and 
WildDash (7.01\%),
and least in
Vistas (0.04\%).
Table \ref{tab:results} summarizes
mIoU performance of the only two 
complete RVC 2020 submissions.
Poor Cityscapes performance of our method
is likely caused by mapping 
all WildDash trains
to only one of the two universal classes
corresponding to on-rail vehicles.
We corrected that in the 32nd   
epoch of training.
Otherwise, our submission prevails
on all datasets except on ADE20k,
which has the smallest images.
This suggests advantage of 
pyramidal fusion 
on large input resolutions.

\begin{table}[htb]
    \centering
    \begin{tabular}{lcc}
    Dataset     & MSeg1080\_RVC     & SN\_RN152pyrx8\_RVC (ours) \\
    \hline
    ADE20K      & \textbf{33.2}     & 31.1 \\ 
    Cityscapes  & \textbf{80.7}     & 74.7 \\ 
    KITTI       & 62.6              & \textbf{63.9} \\ 
    Vistas      & 34.2              & \textbf{40.4} \\ 
    ScanNet     & 48.5              & \textbf{54.6} \\ 
    VIPER       & 40.7              & \textbf{62.5} \\
    WildDash 2  & 35.2              & \textbf{45.4} \\
    \end{tabular}
    \caption{Performance (mIoU)
  of the two RVC'20 submissions.
    }
    \label{tab:results}
\end{table}

\section{Conclusion}

We have described our 
submission to
the semantic segmentation contest
of the Robust Vision Challenge 2020.
Our model outputs dense predictions 
into 193 universal classes,
which requires enormous 
quantity of GPU memory 
during training.
This suggests that memory consumption 
represents a dominant obstacle
towards accurate multi-domain 
dense prediction.
The reported results indicate 
that pyramidal fusion is capable 
to produce competitive performance
in large-scale setups.




\bibliographystyle{splncs}
\bibliography{egbib}

\end{document}


\pagestyle{headings}
\mainmatter

\def\ACCV20SubNumber{973}  

\title{Semi-Supervised Semantic Segmentation \\
    with One-Way Consistency - \\ Supplementary Material} 
\titlerunning{ACCV-20 submission ID \ACCV20SubNumber}
\authorrunning{ACCV-20 submission ID \ACCV20SubNumber}

\author{Anonymous ACCV 2020 submission}
\institute{Paper ID \ACCV20SubNumber}

\maketitle






\section{Additional Experiments}

\subsection{Comparison of One-Way and Two-Way Consistency on CIFAR-10}

We compare the consistency criterion presented in Table~1 of the main paper (1w-cons-tps) with its two-way variants. As noted in subsection~3.3 of the main paper, one-way KL divergence optimization is equivalent to one-way cross entropy optimization.
We thus compare it with two-way cross entropy and two-way KL-divergence in Table~\ref{tab:2w-cifar}. 
We can see that here the two-way KL divergence criterion significantly degrades accuracy with respect to the supervised baseline, which achieves around $81\%$ accuracy. Two-way cross entropy significantly improves upon the baseline, which is consistent with previous work observing benefits of entropy minimization \cite{grandvalet05nips,miyato19pami,xie19arxiv_uda}, and we obtain significantly the best performance with one-way KL divergence.

\begin{table}[htb]
    \centering
    \caption{
      Comparison of one-way and two-way consistency on CIFAR-10 with 4000 labeled examples. The numbers are means of 3 runs. 
    }
    \label{tab:2w-cifar}
    \begin{tabular}{l@{\qquad}c@{\qquad}c@{\qquad}c}
    \hline
    \noalign{\smallskip}
    Method & 
    Accuracy$/\%$ \\
    \noalign{\smallskip}
    \hline
    \noalign{\smallskip}
    1-way KL divergence    & 89.5 \\
    2-way cross entropy    & 84.3 \\
    2-way KL divergence    & 64.6 \\
    \hline
    \end{tabular}
\end{table}

\subsection{Labeled or Unlabeled Coarse?}
In Table \ref{tab:rn34}, we provide an extended version of Table~5 from our manuscript.
We add one additional column, \emph{fine$_\textrm{l}$+coarse$_\textrm{l}$}, 
which corresponds to the standard supervised training on both datasets.
To promote fair comparison, we do not use uniform crop sampling on coarsely labeled images.
That would require leveraging centroids of class instances, which would
not be possible when dealing with unlabeled images.
The results indicate that poor label quality in the coarse subset
yields worse validation accuracy compared to unsupervised training
on coarse images.
We use the single scale SwiftNet \cite{orsic19cvpr} in these experiments.
Experiments with a stronger backbone (ResNet-34)
suggest that regularization effects of semi-supervised training
have greater benefits with higher capacity models.

\begin{table}[htb]
    \centering
    \caption{
      Impact of the coarse split on
      mIoU percentage on Cityscapes val. coarse$_\textrm{u}$ and coarse$_\textrm{l}$ denote whether labels of the coarsely labeled used.
    }
    \label{tab:rn34}
    \begin{tabular}{l@{\qquad}c@{\qquad}c@{\qquad}c}
    \hline
    \noalign{\smallskip}
    Model & 
    fine$_\textrm{l}$ & 
    fine$_\textrm{l}$+coarse$_\textrm{u}$ &
    fine$_\textrm{l}$+coarse$_\textrm{l}$ \\
    \noalign{\smallskip}
    \hline
    \noalign{\smallskip}
    SwiftNet-RN18 & 75.5  &  75.5  & 75.2 \\
    SwiftNet-RN34 & 77.0  &  78.5  & 77.5 \\
    \hline
    \end{tabular}
\end{table}

\subsection{Ensembling Semi-Supervised and Fully Supervised Models}


Table \ref{tab:ens} presents experiments which suggest that
models trained with our semi-supervised loss
operate differently than when they are 
trained with full supervision.
The first row presents an ensemble 
of two ImageNet pre-trained baseline SwiftNet-RN18 models 
as reported in \cite{orsic19cvpr}.
The second row presents an ensemble 
of an ImageNet pre-trained baseline model 
with a randomly initialized model 
trained with our consistency loss.  
The model trained with the consistency loss 
provides a greater contribution to the ensemble
than the baseline model
in spite of being around 1.7 percentage points 
less accurate when performing on its own.

\begin{table}[h]
    \centering
    \caption{
    Results of ensembling different instances of SwiftNet-RN18 on Cityscapes val. We use the parametrization publicly available at \url{https://github.com/orsic/swiftnet}.
    }
    \label{tab:ens}
    \begin{tabular}{l@{\;\;}|@{\;\;}cc@{\;\;}|@{\;\;}c}
    \hline\noalign{\smallskip}
         Models                          & First & Second & Ensemble \\
    \noalign{\smallskip}
    \hline
    \noalign{\smallskip}
         supervised$_1$ + supervised$_2$ \cite{orsic19cvpr}  & 75.35  &  75.43  & 76.45    \\
         supervised$_1$ + 1w-cons-phtps    (ours)   & 75.35  &  73.67  & 76.85    \\
    \end{tabular}
\end{table}

\subsection{Qualitative Results on Cityscapes}

Figure~\ref{fig:cs_examples} illustrates the contribution of semi-supervised training on Cityscapes val. The left column
displays images from Cityscapes val perturbed using $T_{\vec\tau}$.
We trained two SwiftNet-RN18 models, with and
without the proposed consistency term.
The right and middle columns visualize predictions for semi-supervised
and fully-supervised training, respectively.
We notice that our proposed method develops a substantial resilience
to heavy perturbations of the image manifold.

\newcommand{\ssw}{0.33\textwidth}
\begin{figure}[htb]
  \centering
  \includegraphics[width=\ssw]{figs/examples/cs/img_frankfurt_000001_016462.png}\,%
  \includegraphics[width=\ssw]{figs/examples/cs/baseline_frankfurt_000001_016462.png}\,%
  \includegraphics[width=\ssw]{figs/examples/cs/phtps_frankfurt_000001_016462.png}%
  
  \includegraphics[width=\ssw]{figs/examples/cs/img_frankfurt_000001_017101.png}\,%
  \includegraphics[width=\ssw]{figs/examples/cs/baseline_frankfurt_000001_017101.png}\,%
  \includegraphics[width=\ssw]{figs/examples/cs/phtps_frankfurt_000001_017101.png}%
  
  \includegraphics[width=\ssw]{figs/examples/cs/img_frankfurt_000001_076502.png}\,%
  \includegraphics[width=\ssw]{figs/examples/cs/baseline_frankfurt_000001_076502.png}\,%
  \includegraphics[width=\ssw]{figs/examples/cs/phtps_frankfurt_000001_076502.png}%
  
  \includegraphics[width=\ssw]{figs/examples/cs/img_frankfurt_000001_082087.png}\,%
  \includegraphics[width=\ssw]{figs/examples/cs/baseline_frankfurt_000001_082087.png}\,%
  \includegraphics[width=\ssw]{figs/examples/cs/phtps_frankfurt_000001_082087.png}%
  \caption{
    Comparison between the model trained with (right) and
    without (middle) semi-supervised consistency.
    The perturbed image on the left is presented at the model input.
    }
    \label{fig:cs_examples}
\end{figure}

\bibliographystyle{splncs}
\bibliography{egbib}

%% file: math.tex
\let\originalleft\left
\let\originalright\right
\renewcommand{\left}{\mathopen{}\mathclose\bgroup\originalleft}
\renewcommand{\right}{\aftergroup\egroup\originalright}

\usepackage{amsmath}
\usepackage{amssymb}  

\usepackage{mathtools}  

\makeatletter
\newcommand{\spx}[1]{%
	\if\relax\detokenize{#1}\relax
	\expandafter\@gobble
	\else
	\expandafter\@firstofone
	\fi
	{^{#1}}%
}
\makeatother

\newcommand{\genericdel}[4]{%
	\ifcase#3\relax
	\ifx#1.\else#1\fi#4\ifx#2.\else#2\fi\or
	\bigl#1#4\bigr#2\or
	\Bigl#1#4\Bigr#2\or
	\biggl#1#4\biggr#2\or
	\Biggl#1#4\Biggr#2\else
	\left#1#4\right#2\fi
}

\newcommand{\cbr}[2][-1]{\genericdel\{\}{#1}{#2}}
\let\set\cbr

\newcommand{\envert}[2][-1]{\genericdel||{#1}{#2}}

\usepackage{stmaryrd}  

\usepackage[OMLmathsfit]{isomath}  
\usepackage{upgreek}

\DeclareMathAlphabet{\mathbbmsl}{U}{bbm}{m}{sl}
\DeclareMathAlphabet{\mathbbmb}{U}{bbm}{b}{it}
\DeclareMathAlphabet{\mathbbmssit}{U}{bbmss}{m}{it}

\let\set\relax

\newcommand{\set}[1]{\mathbbmsl{#1}}

\newcommand{\nunder}[2][5]{\mathrlap{\mkern\the\numexpr#1/2mu\relax\underline{\phantom{\mathrm{#2}\mkern-#1mu}}}\mathrm{#2}}

\DeclareMathOperator*{\E}{\mathrm{I\kern-.282em E}}
\DeclareMathOperator*{\D}{\mathrm{I\kern-.282em D}}

\let\oldhat\hat
\renewcommand{\hat}[1]{\vphantom{#1}\smash[t]{\oldhat{#1}}}
\let\oldtilde\tilde
\renewcommand{\tilde}[1]{\vphantom{#1}\smash[t]{\oldtilde{#1}}}
\let\oldwidetilde\widetilde
\renewcommand{\widetilde}[1]{\vphantom{#1}\smash[t]{\oldwidetilde{#1}}}

\newenvironment{talign*}
{\let\displaystyle\textstyle\csname align*\endcsname}
{\endalign}

\usepackage{dashbox}%